\PassOptionsToPackage{pdfpagelabels=false}{hyperref} 
\documentclass{IOS-Book-Article}

\usepackage{mathptmx}
\usepackage{soul}\setuldepth{article}

\usepackage{acronym}
\usepackage{xstring}
\usepackage{xcolor}
\usepackage{graphicx}
\usepackage{tikz}
\usepackage[numbers]{natbib}

\usepackage{amsmath}
\usepackage{amssymb}

\usepackage{placeins}

\usepackage{booktabs} % for professional tables
\usepackage{array}
\usepackage{subcaption}
\usepackage{multirow}
\usepackage{hhline}
\usepackage{colortbl}
\usepackage{collcell}
\usepackage{stringstrings}
\usepackage{paralist}
\usepackage[inline]{enumitem}

\newcommand\rurl[1]{%
  \href{http://#1}{\nolinkurl{#1}}%
}

\newcommand*\rot{\rotatebox{90}}%
\newcommand*{\MinNumber}{-1}%
\newcommand*{\MidNumber}{0}%
\newcommand*{\MaxNumber}{1}%

\newcommand{\ApplyGradient}[1]{%
    \getargs[q]{#1}
    \ifdim \narg pt > 0 pt
        \edef\argnumber{\argi}
        \IfDecimal{\argnumber}{
            \ifdim \argnumber pt > \MidNumber pt
                \pgfmathsetmacro{\PercentColor}{max(min(100.0*(\argnumber - \MidNumber)/(\MaxNumber-\MidNumber),100.0),0.00)}
    
                \ifdim \narg pt < 2 pt
                    \edef\x{ \noexpand\cellcolor{cyan!\PercentColor!white} }			% changed colour because green doesn't distinguish well btwn moderately correlated vs highly correlated
                \else
                    \edef\x{ \noexpand\cellcolor{\argii} }
                \fi
                \x \argnumber
            \else
                \pgfmathsetmacro{\PercentColor}{max(min(100.0*(\MidNumber - \argnumber)/(\MidNumber-\MinNumber),100.0),0.00)}
                
                \ifdim \narg pt < 2 pt
                    \edef\x{ \noexpand\cellcolor{green!\PercentColor!green} }
                \else
                    \edef\x{ \noexpand\cellcolor{\argii} }
                \fi
                \x \argnumber
            \fi
        }{\argnumber}
    \else
    \fi
}
\newcolumntype{R}{>{\collectcell\ApplyGradient}r<{\endcollectcell}}

\acrodef{XAI}{explainable AI}
\acrodef{NLP}{natural language processing}

\def\hb{\hbox to 11.5 cm{}}

\begin{document}
\pagestyle{headings} 
\def\thepage{}

\begin{frontmatter}

\title{A Song of (Dis)agreement: Evaluating the Evaluation of Explainable Artificial Intelligence in Natural Language Processing}

\markboth{}{June 2022\hb}

\author[A]{\fnms{Michael } \snm{Neely$^{*\dagger}$}},%
\author[A,B]{\fnms{ Stefan F.} \snm{Schouten$^*$}}{\let\thefootnote\relax\footnote{{* Equal contribution.}}\footnote{{$\dagger$ Now at ASOS.com.}}}
\author[A]{\fnms{ Maurits} \snm{Bleeker}},
and
\author[A]{\fnms{ Ana} \snm{Lucic}}

\runningauthor{}
\address[A]{University of Amsterdam}
\address[B]{Vrije Universiteit Amsterdam}

\begin{abstract}
	There has been significant debate in the NLP community about whether or not attention weights can be used as an \emph{explanation} -- a mechanism for interpreting how important each input token is for a particular prediction. 
	The validity of ``attention as explanation'' has so far been evaluated by computing the rank correlation between attention-based explanations and existing feature attribution explanations using LSTM-based models. 
	In our work, we (i) compare the rank correlation between five more recent feature attribution methods and two attention-based methods, on two types of NLP tasks, and (ii) extend this analysis to also include transformer-based models. 
	We find that attention-based explanations do not correlate strongly with any recent feature attribution methods, regardless of the model or task.
	Furthermore, we find that none of the tested explanations correlate strongly with one another for the transformer-based model, leading us to question the underlying assumption that we should measure the validity of attention-based explanations based on how well they correlate with existing feature attribution explanation methods. 
	After conducting experiments on five datasets using two different models, we argue that the community should stop using rank correlation as an evaluation metric for attention-based explanations. 
	We suggest that researchers and practitioners should instead test various explanation methods and employ a human-in-the-loop process to determine if the explanations align with human intuition for the particular use case at hand. 
\end{abstract}

\end{frontmatter}
\markboth{June 2022\hb}{June 2022\hb}

%!TEX root = ../main.tex

\section{Introduction} \label{sec:introduction}

As machine learning (ML) models are increasingly used in hybrid settings to make consequential decisions for humans, criteria for plausible and faithful explanations of their predictions remain speculative \citep{lipton2016mythos, jacovi-goldberg-2020-towards}.
Although there are many possible explanations for a model's decision, only those faithful to both the model's reasoning process and to human stakeholders are desirable \cite{jacovi-goldberg-2020-towards}. 
The rest are irrelevant in the best case and harmful in the worst, particularly in critical domains such as law \cite{Kehl2017AlgorithmsIT}, finance \cite{grath_interpretable_2018}, and medicine \cite{caruana2015intelligble}.

Content moderation is a use case where explanations can help domain experts as part of a hybrid human-in-the-loop system. 
Consider an ML model that predicts whether or not a post on a social media contains misinformation: when a post is automatically removed by the model and the user who created it appeals its removal, content moderators need to read through the entire post to identify why it was flagged as containing misinformation \cite{halevy_preserving_2022}. 
Including explanations that identify which parts of the post are problematic can help content moderators decide if the model was correctly flagging the post. 

ML practitioners frequently explain models by calculating each input's contribution toward an individual prediction.
Additivity --- treating all contributions as independent and quantifiable --- is a common simplifying assumption. 
In this work, we focus on such additive explanation methods and refer to them as \emph{feature attribution methods}. 
We denote the contribution of each input toward the model's decision as its \emph{importance}. We say that two different feature attribution methods \emph{agree} if there is a strong correlation between their computed rankings of input importance.

% Introducing Attention-Based Explanations
The attention mechanism \cite{bahdanau2014neural} in natural language processing (NLP) is a popular, albeit less rigorously motivated, way of obtaining explanations. Because the mechanism produces context vectors from which decoders can soft-search for prediction-relevant information, the weights assigned to inputs intuitively serve as proxies for their overall contribution towards a decision. Weights are often visualized as heatmaps over sequences \cite[e.g.,][]{pmlr-v37-xuc15,li-etal-2016-visualizing}, which can be particularly persuasive when examples are (unintentionally) cherry-picked to fit a narrative. 
We define an attention-based explanation as a vector of attention weights that, similar to a feature attribution method, can be treated as a ranking of importance.

In their critique of attention-based explanations for NLP, \citet{jain-wallace-2019-attention} argue that faithful attention-based explanations must be highly \emph{agreeable}.\footnote{\citet{ethayarajh-jurafsky-2021-attention} use the term \emph{consistent}.} That is, their generated rankings of input importance must correlate with existing feature attribution methods. 
Following \citet{jain-wallace-2019-attention} and their claim that ``attention is not explanation'', several recent papers have presented an increased agreement with a small set of feature attribution methods as evidence for their proposed method's ability to improve the faithfulness of the attention mechanism. For example, \citet{mohankumar-etal-2020-towards} show that minimizing hidden state conicity in a BiLSTM improves the Pearson correlation of attention weights with Integrated Gradients \cite{intgrad} attributions. As the popularity of \emph{agreement as evaluation} grows \citep{meister-etal-2021-sparse,abnar-zuidema-2020-quantifying}, we believe it is important to investigate diagnostic capacity of agreement as a metric by examining (i) more recent feature attribution methods, and (ii) more complex transformer-based models. 

Under the paradigm of \emph{agreement as evaluation}, new explanation methods (i.e., attention-based) are compared to established explanation method(s) (i.e., feature attributions). 
However, can any one explanation method act as the standard against which other explanation methods are evaluated?
Explanations are task-, model-, and context-specific \cite{doshi-velez_towards_2017}, and the performance of explanation methods depends on the particular diagnostic tests considered \citep{deyoung-etal-2020-eraser,Robnik-sikonja2018}. 

In this work, our main research question is: \emph{How well do attention-based explanations correlate with recent feature attribution methods for NLP tasks?} Specifically, we want to investigate: 
\begin{itemize}
	\item \textbf{RQ1:} Does the correlation depend on the model architecture (transformer- vs. LSTM-based)?
	\item \textbf{RQ2:} Does the correlation depend on the nature of the classification task (single- and pair-sequence)? 
\end{itemize}

We investigate the following feature attribution methods: {LIME} \cite{lime}, {Integrated Gradients} \cite{intgrad}, {DeepLIFT} \cite{deeplift}, and two versions of SHAP \cite{shap}: 
{Grad-SHAP} (based on Integrated Gradients) and {Deep-SHAP} (based on DeepLIFT). 
We observe low agreement between attention-based explanations and feature attribution methods, across both models and both tasks. 
We also observe low agreement across all explanation methods for the transformer-based model and for pair-sequence tasks. 
We use this empirical evidence, along with our theoretical objections, to argue that practitioners should refrain from evaluating attention-based explanations based on their agreement with feature attribution methods.

%!TEX root = ../main.tex

\section{Related Work}

\citet{jain-wallace-2019-attention} introduced the \emph{agreement as evaluation} paradigm by comparing attention-based explanations with simple feature attribution methods using a bidirectional LSTM \cite{lstm} on single- and pair-sequence classification tasks. 
They conclude that ``attention is not explanation'' due to the weak correlation between the rankings of input token importance obtained from attention weights and those obtained from two elementary feature attribution methods: (i) input $\times$ gradient \citep{inputxgradients1, inputxgradients2}, and (ii) leave-one-out \cite{li2017understanding}. 
In our work, we test the generalizability of agreement as an evaluation metric by (i) testing on a more comprehensive set of feature attribution methods (see Section~\ref{section:xai-methods}), and (ii) testing on a transformer-based \cite{vaswani2017attention} model on the same types of tasks (see Section~\ref{section:models}). 

The influential critique by \citet{jain-wallace-2019-attention} has sparked an ongoing debate about whether or not attention is explanation \citep{wiegreffe-pinter-2019-attention, serrano-smith-2019-attention, grimsley-etal-2020-attention}. 
More recently, \citet{bastings-filippova-2020-elephant} have questioned the notion of ``attention as explanation'' as a whole, and suggest that in order to explain ML model predictions, the community should rely on methods that are explicitly created for this purpose (i.e., feature attributions), instead of seeking explanations from attention mechanisms. 
In our work, we do not aim to take a position on the ``is attention explanation'' debate, but rather investigate the hypothesis that in order for attention mechanisms to be considered ``explanations'', they must correlate with existing feature attribution methods. This is an underlying assumption of not only the work by \citet{jain-wallace-2019-attention}, but also that of \citet{meister-etal-2021-sparse}, who show that inducing sparsity in the attention distribution decreases agreement with feature attribution methods, and \citet{abnar-zuidema-2020-quantifying}, who demonstrate their \emph{attention-flow} algorithm improves the correlation with attributions based on feature ablation. 

\citet{atanasova-etal-2020-diagnostic} introduce a series of diagnostic tests to evaluate feature attribution methods for text classification. 
They show that the performance of feature attribution methods, measured by using these diagnostic tests, largely depends on the model and task considered, but note that gradient-based methods tend to perform the best. 
Similar to their work, we also compare and evaluate feature attribution methods, but only to investigate the suitability of \textit{agreement as evaluation}, not to determine a winning explanation method given several diagnostic tests. 
In contrast, we evaluate five feature attribution methods on agreement as evaluation based on only the rank correlation using Kendall's-$\tau$, following \citet{jain-wallace-2019-attention}.

\citet{prasad-etal-2021-extent} define three alignment metrics that quantify how well human-annotated natural language explanations align with the explanations generated by the Integrated Gradients method \citep{intgrad}. 
They find that the BERT \cite{devlin-etal-2019-bert} model has the highest alignment with human-annotated explanations. 
However, unlike our work, this work focuses exclusively on using transformer-based models and Integrated Gradients, and does not question the \emph{agreement as evaluation} paradigm for feature attribution methods. 

\citet{ding-koehn-2021-evaluating} introduce a human-annotated benchmark to evaluate feature attribution methods for NLP models. 
They test two attribution methods on three types of NLP models and find that explanations from feature attribution methods are sensitive to changes in model configuration. 
Similarly, we test five attribution methods on two types of NLP models, but our focus is on investigating the \emph{agreement as evaluation} paradigm, whereas \citet{ding-koehn-2021-evaluating} focus on investigating the correlation between feature attribution methods and a human-annotated benchmark. 
We are unable to incorporate the benchmark proposed by \citet{ding-koehn-2021-evaluating} in our work because the human annotations are not present for every single token, and therefore we cannot turn them into a ranking. 

% New publications since our Workshop paper
This work builds off our prior ICML workshop paper \cite{neely2021order}, which several other papers have extended. \citet{feldhus-etal-2021-thermostat} introduce a software package to analyze instance-wise explanations for popular NLP models and tasks and, in doing so, partially reproduce one of our experiments. \citet{krishna_disagreement_2022} formally define and highlight the importance of the "disagreement problem" between feature attribution methods, which they find is a constant frustration for ML practitioners. They introduce metrics to capture the disagreement between top-k features and ``features of interest'' (e.g., those selected by an end-user) and find considerable disagreement between feature attribution methods for tabular, text, and image data on both real-world and research datasets.

%!TEX root = ../main.tex

\section{Explainability in NLP} \label{sec:method}
In this work we investigate two types of explanations for NLP tasks: (i) those from recent feature attribution methods in the \ac{XAI} literature and (ii) those based on attention scores. We evaluate on transformer- and LSTM-based models (\textbf{RQ1}), for both single- and pair-sequence tasks (\textbf{RQ2}). 
Following \citet{jain-wallace-2019-attention}, we define an \emph{explanation} of an input token sequence as a vector comprised of an importance score for each token. 
This vector can be used to rank the tokens by their importance score.

\subsection{Explanations from Feature Attributions} \label{ssec:recent_methods}
\label{section:xai-methods}

In our experiments, we focus on five recent feature attribution methods: {LIME} \cite{lime}, {Integrated Gradients} \cite{intgrad}, {DeepLIFT} \cite{deeplift}, and two versions of SHAP \cite{shap}: 
{Grad-SHAP} and {Deep-SHAP}:

\begin{itemize}
	\item LIME \cite{lime} produces locally faithful explanations by learning an interpretable (e.g., linear) model from samples weighted by their proximity to the original instance.
	\item Integrated Gradients \cite{intgrad} calculates input feature attributions by accumulating the gradients obtained from the model along the straight-line path from a baseline to the original instance.
	\item DeepLIFT \cite{deeplift} also produces input feature attributions using the gradients, but it assigns scores based on the difference between a reference activation and the activation of the original instance. This allows the calculated contributions to remain non-zero even when the gradients are zero. 
	\item SHAP \cite{shap} identifies a unique solution for the contribution of each input toward the prediction. Since this is computationally expensive, \citet{shap} propose approximations based on existing methods: 
	\begin{itemize}
		\item[$\circ$] Grad-SHAP (based on Integrated Gradients). 
		\item[$\circ$] Deep-SHAP (based on DeepLIFT).
	\end{itemize}
\end{itemize} 

\subsection{Explanations from Attention Mechanisms} \label{ssec:attn_methods}

Given an input sequence of tokens $\textbf{S} = (t_{1}, ..., t_{n})$, we define an \emph{attention-based explanation} as an assignment of attention weights $\boldsymbol{\alpha} \in \mathbb{R}^{n}$ to the tokens in $\textbf{S}$ \citep{jain-wallace-2019-attention}. 
Since the dimensionality of $\boldsymbol{\alpha}$ is architecture-dependent, it may be necessary to filter or aggregate the weights. In our experiments, this is only relevant for the self-attention mechanism in the transformer-based model we consider below (see Section~\ref{section:models}). 

Previous analyses at the attention head level implicitly assume that contextual word embeddings remain tied to their corresponding tokens across self-attention layers \citep{baan2019understanding, clark-etal-2019-bert}. 
This assumption may not hold in transformer-based models since information mixes across layers \cite{brunner2019identifiability}. 
Therefore, we use the \emph{attention rollout} \cite{abnar-zuidema-2020-quantifying} method --- which assumes the identities of tokens are linearly combined through the self-attention layers based exclusively on attention weights --- to calculate post-hoc, token-level importance scores.%
\footnote{We also experimented with \textit{attention flow} \cite{abnar-zuidema-2020-quantifying}, see Appendix \ref{app:attention_flow}.}
Following \citet{abnar-zuidema-2020-quantifying}, we use the scores calculated for the last layer's [CLS] token, resulting in a final vector $\boldsymbol{\alpha} \in \mathbb{R}^{n}$ at the time of evaluation.

Recurrent models similarly suffer from issues of identifiability. In LSTM-based models, attention is computed over hidden representations across timesteps, which does not provide faithful token-level importance scores. Approaches that trace explanations back to individual timesteps \cite{bento_timeshap_2021} or input tokens \cite{tutek-snajder-2020-staying} are only just emerging. 
Therefore, we analyze the raw attention weights for the LSTM-based model we consider below (see Section~\ref{section:models}).

\subsection{Agreement between Explanations}

Following \citet{jain-wallace-2019-attention}, we measure \emph{agreement} between the explanation methods as the mean Kendall-$\tau$ correlation \cite{kendall-tau} between the ranked importance scores of all input tokens, across all examples.
% \footnote{Note that Spearman and Pearson correlations were also calculated which yielded very similar results that do not conflict with any of our conclusions.}
\footnote{We also calculated Spearman and Pearson correlations and obtained similar results.}
The Kendall-$\tau$ correlation is a widely used metric for comparing ranked lists; it measures the correlation between two ranked lists based on discordant pairs between the lists. Two items are considered \emph{discordant} if they are ranked differently on the two lists. 
The Kendall-$\tau$ correlation can take on values in the $[-1, 1]$ interval, where negative values imply the rankings are negatively correlated while positive values imply the rankings are positively correlated \cite{kendall-tau}. 

%!TEX root = ../main.tex
\section{Experimental Setup} \label{sec:experiments}

\subsection{Datasets} \label{sec:datasets}

We evaluate on two types of NLP classification tasks: 
\begin{enumerate*}[label=(\roman*)]
	\item single-sequence, and
	\item pair-sequence. 
\end{enumerate*}
For the single-sequence task, we perform binary sentiment classification on the Stanford Sentiment Treebank (SST-2) \cite{socher-etal-2013-parsing} and the IMDb Large Movie Reviews Corpus (IMDb) \cite{maas-etal-2011-learning}. 
We use identical splits and pre-processing as \citet{jain-wallace-2019-attention}, but also remove sequences longer than 240 tokens for faster attribution calculation. This leaves us with roughly 78\% of the original instances in the IMDb dataset.

\begin{itemize}
	\item The SST-2 dataset consists of single sentences extracted from movie reviews and is used for binary sentiment analysis: predicting whether the review is positive or negative. 
	\item The IMDb dataset consists of movie reviews as well, but contains longer sequences compared to SST-2. It is also used for binary sentiment analysis.
\end{itemize}

For the pair-sequence task, we examine natural language inference and understanding with the Multi Natural Language Inference (MNLI) corpus \cite{williams-etal-2020-predicting}, the Stanford Natural Language Inference (SNLI) corpus \cite{bowman-etal-2015-large}, and the Quora Question Pairs dataset \cite{quora}. 

\begin{itemize}
	\item The MNLI dataset contains sentence pairs for a textual entailment task: given a pair of sentences, we want to predict whether or not one sentence implies the other. Since MNLI has no publicly available test set, we use the English subset of the XNLI \cite{conneau2018xnli} test set. 
	\item The SNLI dataset~\cite{bowman-etal-2015-large} contains pairs of sentences and is used for the textual entailment task, similar to MNLI. 
	\item The Quora Question Pairs dataset \cite{quora} contains pairs of questions from the Quora website, where the task is to classify if the two questions in a pair are duplicates or not. We use a custom split (80/10/10) for the Quora dataset, removing pairs with a combined count of 200 or more tokens (leaving 99.99\% of the original instances).
\end{itemize}

\subsection{Models}
\label{section:models} 

We test two types of NLP models: 
\begin{enumerate*}[label=(\roman*)]
	\item transformer-based, and
	\item LSTM-based. 
\end{enumerate*}
The transformer-based model \cite{vaswani2017attention} relies on a self-attention mechanism to calculate representations for tokens.  
For each layer in the transformer network, the representation of each token is updated by computing a weighted sum over all tokens represented in the entire sequence, together with a non-linear feedforward transformation. 
The weight value of each token is determined using self-attention, which computes a similarity score for each pair of tokens in the sequence. 
For every token in every layer of the network, an attention layer is used. 
As a result, it is non-trivial to aggregate all the importance values per layer into one interpretable importance score per token. 

In our work, we follow \citet{abnar-zuidema-2020-quantifying} by using the attention rollout scores for [CLS] tokens. 
For the transformer-based model, we fine-tune the lighter, pre-trained DistilBERT variant \cite{sanh_distilbert_2019} instead of the full BERT model \cite{devlin-etal-2019-bert} to reduce the computational overhead and ecological footprint. For classification, we add a linear layer on top of the pooled output. We concatenate pair-sequences with a [SEP] token.

Unlike a transformer-based model, an LSTM-based model processes input tokens in sequential order. 
For each token, $t \in \{1, \dots, n\}$, in the sequence, the global representation of the sequence, $\textbf{h}_{t}$, is updated using a non-linear transformation with an embedded representation of 
\begin{inparaenum}[(i)]
	\item the current token in the sequence, $t_{n}$, and 
	\item the previous global representation of the sequence, $\textbf{h}_{t-1}$, as input.
\end{inparaenum}
A bidirectional LSTM (Bi-LSTM) uses two stacked LSTMs to process tokens in both directions of the sequence. 

For the Bi-LSTM, we use the same single-layered bidirectional encoder with the query-less additive $(tanh)$ attention and linear feedforward decoder as in the work of \citet{jain-wallace-2019-attention}. 
As a result, the attention weight for each token in the sequence is solely based on its representation $\textbf{h}_{t}$, and not on a query item.
In pair-sequence tasks, we embed, encode, and induce attention over each sequence separately. 
The decoder predicts the label from the concatenation of 
\begin{inparaenum}[(i)]
	\item both context vectors $c_{1}$ and $c_{2}$, 
	\item their absolute difference $|c_{1} - c_{2}|$, and
	\item their element-wise product $c_{1} \odot c_{2}$.
\end{inparaenum}

\subsection{Training the models}

We train three independently-seeded instances of each of the models described in Section~\ref{section:models} using the AllenNLP framework \cite{gardner-etal-2018-allennlp}, each for a maximum of 40 epochs. We use a patience value of 5 epochs for early stopping. 
For DistilBERT, we fine-tune the standard ``base-uncased'' weights available in the HuggingFace library \cite{Wolf2019HuggingFacesTS} with the AdamW \cite{loshchilov2019decoupled} optimizer. 
For the BiLSTM, we follow \citet{jain-wallace-2019-attention} and select a 128-dimensional encoder hidden state with a 300-dimensional embedding layer. We tune pre-trained FastText embeddings \cite{bojanowski-etal-2017-enriching} and optimize with the AMSGrad variant \cite{Tran_2019} of Adam \cite{kingma_dp_adam_2015}. Appendix \ref{app:performance} details model performance and indicates both the BiLSTM and DistillBERT are sufficiently accurate for our analysis.

\subsection{Explaining the models}

We leverage the Captum\footnote{\url{https://github.com/pytorch/captum}} implementations of LIME, Integrated Gradients, DeepLIFT, Grad-SHAP, and Deep-SHAP, and use the padding token as a baseline where applicable. For LIME, we mask tokens as features and use 1000 samples to train the interpretable models. 
We apply our feature attribution methods to the predictions of each independently-seeded model for 500 instances randomly sampled from each test set.

Our code is publicly available.\footnote{\url{https://github.com/sfschouten/court-of-xai}} Refer to Appendix \ref{app:reproducibility} for more information on reproducing our experiments.

%!TEX root = ../main.tex

\section{Results}\label{sec:results}

In this section, we answer our main research question: \emph{How well do attention-based explanations correlate with recent feature attribution methods for NLP tasks?} 
In general, we find that attention-based explanations do not correlate strongly with feature attribution methods, with some exceptions (see Section~\ref{sec:results2} and~\ref{sec:results3}). 

\subsection{RQ1: Does the correlation depend on the model architecture?}
\label{sec:results1}
Tables~\ref{tab:distilbert_results} and~\ref{tab:bilstm_results} display the average\footnote{Across the 3 model instances, randomly selecting 500 instances from the test set using the training seed.}  Kendall-$\tau$ correlations between the explanation methods for the DistilBERT and BiLSTM models, respectively.\footnote{See Appendix \ref{app:results-with-std-dev} for the same tables including the standard deviation.}
A stronger correlation (i.e., agreement) is indicated by a darker blue colour in the table cell. 
In general, we see that the agreement between explanation methods is substantially lower for the DistilBERT model than for the BiLSTM model. 

\begin{table}[ht!]
    \centering
    \caption{Mean Kendall-$\tau$ between the tested explanation methods for the DistillBERT model. A darker color indicates a stronger correlation between the compared explanation methods. Attn Roll refers to Attention Rollout.}
    \label{tab:distilbert_results}%
    \resizebox{0.7\columnwidth}{!}{%
        \begin{tabular}{clRRRRRR}
            \toprule
               &  & \multicolumn{1}{c}{Attn Roll}   & \multicolumn{1}{c}{LIME} & \multicolumn{1}{c}{Int-Grad} & \multicolumn{1}{c}{DeepLIFT} & \multicolumn{1}{c}{Grad-SHAP} & \multicolumn{1}{c}{Deep-SHAP}            \\
            \midrule
            
            \multirow{5}{*}{\rot{Attn Roll}}
            	& IMDb & 1    & .1259  & .1818  & .2516     & .1432 & .2303   \\
            	& SST-2  & 1 & .1359  & .0511  & .1328     & .0737 & .1291   \\
                & MNLI  & 1  & .2678  & .1891  & .2432     & .1905 & .2067    \\
                & Quora & 1  & .1622  & .0574  & .2267     & .0518 & .2257   \\
                & SNLI  & 1  & .1434  & .1645  & .2214     & .1600 & .1796   \\
      \midrule
            \multirow{5}{*}{\rot{LIME}}
            	& IMDb   & &   1     & .1050   & .0696    & .0929 & .0655   \\
            	&SST-2 &   &  1      & .2861   & .0618    & .2414 & .0499   \\
                & MNLI  &   &   1     & .1794   & .1526    & .1592 & .1205 \\
                & Quora &  &   1     & .1407   & .0032    & .1144 & .0095   \\
                & SNLI  &  &   1     & .1529   & .0925    & .1104 & .0593   \\
     \midrule
            \multirow{5}{*}{\rot{Int-Grad}}
            	& IMDb  &  &        & 1 & .1433    & .5495  & .1246   \\
            	& SST-2 &   &        & 1 & .0498    & .4987  & .0381   \\
                & MNLI &   &        & 1 & .2153    & .4780  & .1708   \\
                & Quora &  &        & 1 & .0625    & .4674  & .0529   \\
                & SNLI  &  &        & 1& .0955    & .3932  & .0700   \\
      \midrule
            \multirow{5}{*}{\rot{DeepLIFT}}
            	& IMDb  &   &        &&  1   &   .1306   & .4830   \\
            	& SST-2 &   &        &&  1   &   .0522   & .4514   \\
                & MNLI  &  &        & &  1   &   .2324   & .4985   \\
                & Quora &   &        & & 1    &   .0637   & .5951   \\
                & SNLI  &  &        &&  1   &   .1181   & .5554   \\
      \midrule
            \multirow{5}{*}{\rot{Grad-SHAP}}
            	& IMDb  &  &        &         &     &   1    & .1093   \\
            	& SST-2 &  &        &         &     &   1    & .0419   \\
                & MNLI  &  &        &         &     &   1    & .1752   \\
                & Quora &  &        &         &     &  1     & .0535   \\
                & SNLI  &   &        &         &     &  1     & .0851   \\
        \midrule
	          	\multirow{5}{*}{\rot{Deep-SHAP}}
	          	& IMDb  &  &        &         &     &       & 1   \\
	          	& SST-2 &  &        &         &     &       & 1   \\
	          	& MNLI  &  &        &         &     &       &  1   \\
	          	& Quora &  &        &         &     &       & 1   \\
	          	& SNLI  &   &        &         &     &       & 1  \\
		        \bottomrule
        \end{tabular}}

\end{table}

For the DistilBERT model, we observe a weak correlation across all explanations -- almost none of the explanation methods we test, whether they are attention-based or feature attributions, correlate strongly with one another. 
There are two exceptions: (i) Integrated Gradients moderately correlates with Grad-SHAP, and (ii) DeepLIFT moderately correlates with Deep-SHAP. 
However, this is unsurprising since the implementation of Grad-SHAP is based on Integrated Gradients, and the implementation of Deep-SHAP is based on DeepLIFT. 
In contrast, it is surprising to see the lack of correlation between Grad-SHAP and Deep-SHAP for the DistilBERT model, given that they are different implementations of the same algorithm, SHAP \cite{shap}. 

\begin{table}[ht!]
    \centering
    \caption{Mean Kendall-$\tau$ between the tested explanation methods for the BiLSTM. A darker color indicates a stronger correlation between the compared explanation methods.}
    \label{tab:bilstm_results}
     \resizebox{0.7\columnwidth}{!}{%
        \begin{tabular}{clRRRRRR}
            \toprule
            &  & \multicolumn{1}{c}{Attn Weights}  & \multicolumn{1}{c}{LIME} & \multicolumn{1}{c}{Int-Grad} & \multicolumn{1}{c}{DeepLIFT} & \multicolumn{1}{c}{Grad-SHAP}  & \multicolumn{1}{c}{Deep-SHAP} \\
           \midrule
            \multirow{5}{*}{\rot{Attn Weights}}
            	& IMDb & 1   & .2014  & .2188  & .2494     & .2209 & .2309   \\
            	& SST-2  &  1 & .1326  & .1093  & .1372     & .1101 & .1400   \\
                & MNLI   &  1& .1958  & .2523  & .2549     & .2473 & .2370   \\
                & Quora  & 1 & .0363  & .0143  & .0894     & .0182 & .1017   \\
                & SNLI   & 1  & .2198  & .2566  & .3158     & .2517 & .2938   \\
           \midrule
            \multirow{5}{*}{\rot{LIME}}
             	& IMDb    & &    1    & .6538   & .5854    & .6486 & .5584   \\
            	& SST-2  & &  1      & .4968   & .4734    & .4962 & .4422   \\
                & MNLI   & &   1     & .3281   & .2444    & .3187 & .2269   \\
                & Quora   & &    1    &  .2099  & .1900    & .2037 & .1670   \\
                & SNLI    & &    1    & .2673   & .1676    & .2481 & .1566   \\
          \midrule
            \multirow{5}{*}{\rot{Int-Grad}}
            	& IMDb  &  &    &  1   & .7331   & .9409  & .6994   \\
            	& SST-2 & &    &  1   & .8683   & .9707  & .8063   \\
                & MNLI & &    &   1  & .4984   & .8138  & .4021   \\
                & Quora & &    &   1  & .2906   & .7420  & .2290   \\
                & SNLI &  &    &   1  & .2461   & .6535  & .2165   \\
          \midrule
            \multirow{5}{*}{\rot{DeepLIFT}}
            	& IMDb  &    &        &&   1  &   .7378    & .8593   \\
            	& SST-2 &   &        &&   1  &   .8682    & .8729   \\
                & MNLI   & &        &&   1  &   .4987    & .6208   \\
                & Quora &  &        &&  1   &   .3158    & .6179   \\
                & SNLI  &  &        &&  1   &   .2557    & .5791   \\
          \midrule
            \multirow{5}{*}{\rot{Grad-SHAP}}
            	& IMDb &    &        &         &     &    1   & .7021   \\
            	& SST-2 &  &        &         &     &    1   & .8056   \\
                & MNLI  &  &        &         &     &    1   & .4015   \\
                & Quora &  &        &         &     &  1     & .2433   \\
                & SNLI &   &        &         &     &    1   & .2219   \\
          \midrule
             \multirow{5}{*}{\rot{Deep-SHAP}}
	            & IMDb &    &        &         &     &     & 1   \\
	            & SST-2 &  &        &         &     &      & 1  \\
	            & MNLI  &  &        &         &     &      &  1   \\
	            & Quora &  &        &         &     &       & 1   \\
	            & SNLI &   &        &         &     &       &  1   \\
          \bottomrule
            \end{tabular}}
\end{table}

For the BiLSTM model, we observe a weak correlation between attention-based explanations and feature attribution explanations (i.e., the first row of Table~\ref{tab:bilstm_results}). 
Similar to the DistilBERT model, we see strong correlations for the methods that have similar underlying implementations. 
We also see some strong correlation between feature attribution methods, especially for single-sequence tasks (see Section~\ref{sec:results2}). 

Overall, we conclude that the correlation between explanation methods depends on the model architecture, which answers \textbf{RQ1}. 
In general, the correlation is weaker for the DistilBERT model than for the BiLSTM. 
However, the overall agreement between attention-based explanations and feature attribution explanations is weak, regardless of the model architecture.  

\subsection{RQ2: Does the correlation depend on the nature of the classification task?} 
\label{sec:results2}
To investigate \textbf{RQ2}, we examine five different datasets corresponding to two different tasks: IMDb and SST-2 are single-sequence tasks, while MNLI, Quora, and SNLI are pair-sequence tasks and show the results in Tables~\ref{tab:distilbert_results} and~\ref{tab:bilstm_results}. 

For the BiLSTM model, the results are clear: there is a substantially stronger correlation among explanations for single-sequence tasks than for pair-sequence tasks. 
For the DistilBERT model however, we do not see such a clear distinction -- there is, at best, a weak correlation across all explanation methods, regardless of the nature of the task. 

Regarding \textbf{RQ2}, we conclude that explanation methods agree with each other more on single-sequence tasks than pair-sequence tasks. This relationship holds especially for the BiLSTM model compared to the DistilBERT model.

\subsection{Explanations that strongly correlate}
\label{sec:results3}

Given that some of the feature attribution methods are inherently related, we would expect to see stronger degrees of correlation between them: 
\begin{enumerate*}[label=(\roman*)]
	\item  Grad-SHAP and Deep-SHAP are two different versions of the SHAP explanation method \cite{shap},
	\item  Grad-SHAP relies on the Integrated Gradients implementation to compute the token importances, and 
	\item Deep-SHAP relies on DeepLIFT.
\end{enumerate*}
We see some strong correlations for (ii) and (iii) between these methods for both models. 
These correlations are especially strong for the single-sequence tasks on the BiLSTM. 
However, we see remarkably weak correlations for (i) on DistilBERT, and for the BiLSTM on the pair-sequence tasks.

%!TEX root = ../main.tex

\section{Discussion} \label{sec:discussion}
Based on the results in Section~\ref{sec:results}, we argue that rank correlation with existing feature attribution methods is not an appropriate measure for evaluating attention-based explanations. 
In this section, we detail three main reasons for this: 
\begin{enumerate*}[label=(\roman*)]
	\item the general lack of correlation across all explanation methods, especially for transformer-based models,
	\item the fact that similar explanations do not always result in correlated rankings, and 
	\item the lack of justification for the existence of one ``ideal'' explanation, which is a fundamental assumption of the \emph{agreement as evaluation} paradigm. 
\end{enumerate*}

\subsection{Lack of correlation between explanation methods.}
In Section~\ref{sec:results}, we have shown that there is a low degree of correlation between explanation methods, especially for the transformer-based model.\footnote{Although there are some examples of stronger correlations, these are either on a very specific combination of task and model (i.e., single sequence tasks on the BiLSTM), or due to similarities in the underlying implementation of the explanation methods. } Similar conclusions are observed in the work of \citep{feldhus-etal-2021-thermostat} and \citep{krishna_disagreement_2022}.
This makes it challenging to justify the expectation that in order for attention-based explanations to be valid, they should correlate with existing feature attribution methods. 
If none of the existing feature attribution methods correlate with one another (as is the case for the transformer-based model), should we expect attention-based explanations to correlate with them?
Therefore, we argue against the use of rank correlation as an ``off-the-shelf'' tool to evaluate attention-based explanations. 

A common critique of using Kendall-$\tau$ is that expecting the full rankings of token importance to correlate is unrealistic. In their original paper, \citet{jain-wallace-2019-attention} theorized that top-$k$ token comparison would reduce noise, improve correlation, and more closely align with the end user's interest in the most salient features. However, formulating a systematic method for calculating $k$ across various models, tasks, and datasets is difficult in practice and yields mixed results. \citet{treviso-martins-2020-explanation} show a dynamic selection of $k$ with a sparse attention mechanism more effectively conveys justifications of decisions than fixed values of $k$ (e.g., 5 or 10 tokens). In contrast, \citet{krishna_disagreement_2022} show a widespread disagreement of feature attribution methods across domains regardless of whether $k$ is fixed or dynamic (e.g., the top $X$\% of features or top $X$\% of tokens based on average sentence length).

\subsection{Similar explanations do not imply strongly correlated rankings}
\begin{figure}
	\centering
	\begin{subfigure}{.48\textwidth}
		\centering
		\includegraphics[width=1\linewidth]{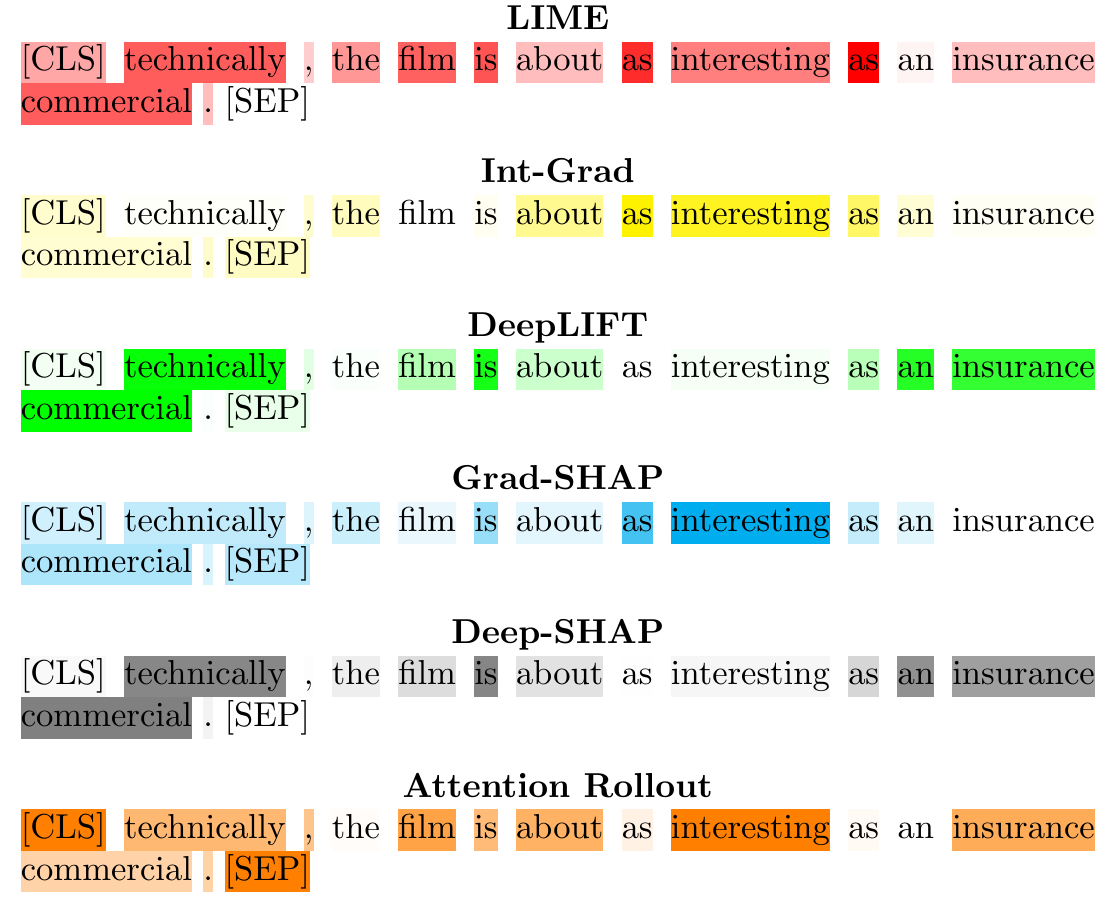}
		\caption{Sentiment analysis example from the SST-2 dataset. The average Kendall-$\tau$ correlation across all methods for this example is $0.01$. }
%		\vspace*{-16mm}
		\label{fig:distill-ss}
	\end{subfigure}\hspace{1mm}
	\begin{subfigure}{.5\textwidth}
		\centering
		\includegraphics[width=1\linewidth]{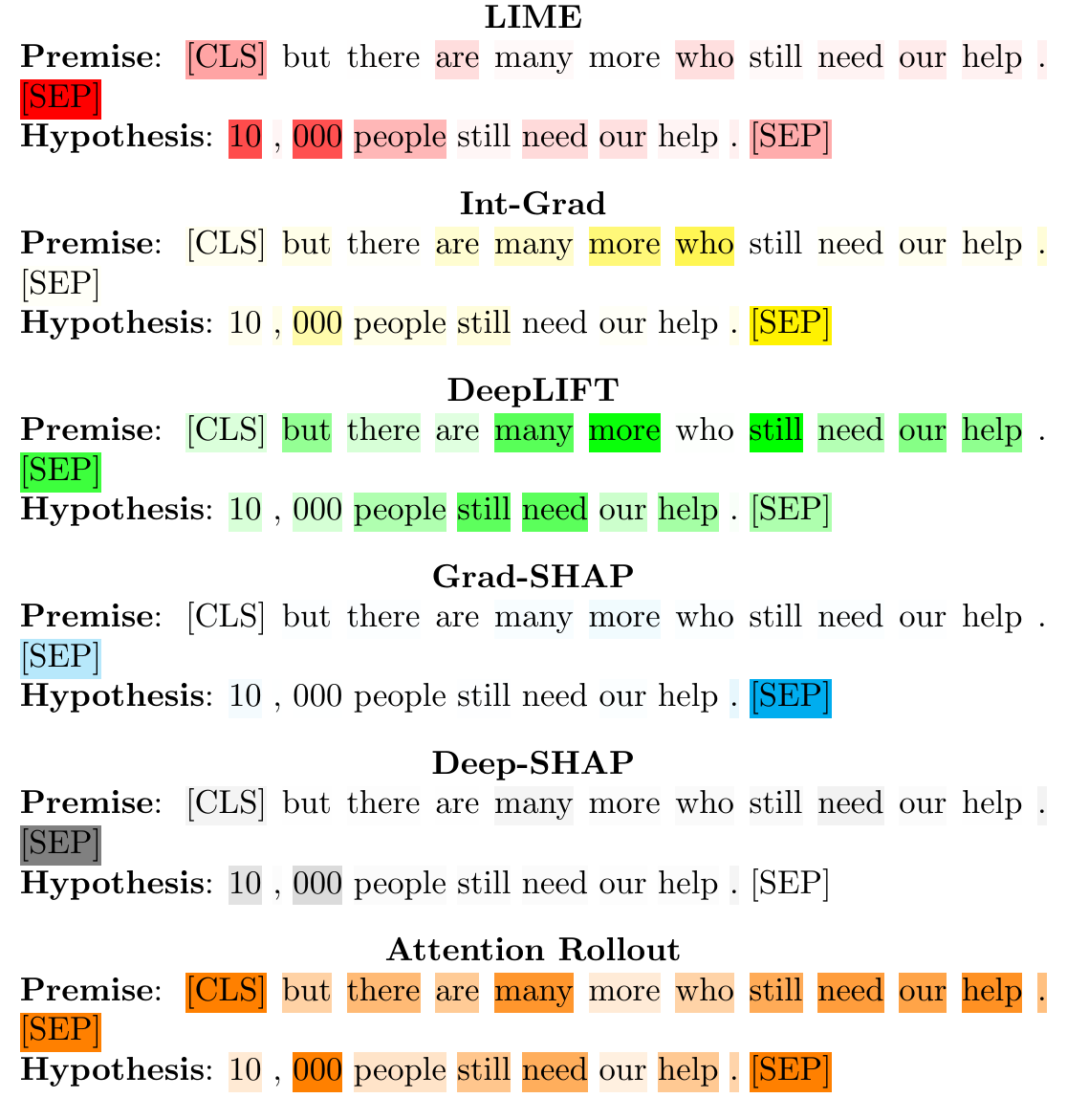}
		\caption{Textual entailment example from the MNLI dataset.The average Kendall-$\tau$ correlation across all methods for this example is $0.05$. }
		\label{fig:distill-ps}
	\end{subfigure}
	\caption{Examples of explanations for the transformer-based model. The brighter the color, the higher the attribution value}
	\label{fig:test}
\end{figure}

In order to investigate what weak correlation looks like in practice, we apply a heatmap of importance scores for each of the tested explanation methods. 
The darker the color, the more important the token is w.r.t. the model prediction. 
Figure~\ref{fig:distill-ss} shows a sentiment analysis example using the transformer-based model on the SST-2 dataset. 
Upon first glance, the visualizations in Figure~\ref{fig:distill-ss} look relatively similar: the methods are all highlighting roughly the same words -- almost all methods (except for Int-Grad) highlight the word `technically' as important. 
However, the average Kendall-$\tau$ correlation across all methods for this example is very weak -- only $0.01$. This highlights the problem of evaluating explanations by measuring the correlation between two ranked lists of importance scores: two explanation methods can roughly indicate the same token(s) as important, but when similar tokens do not end up in similar positions in the two rankings, the overall agreement between the two methods can still be low.

\subsection{Is there one ideal explanation?}
The \emph{agreement as evaluation} paradigm implicitly assumes the existence of a single ``ideal'' explanation (i.e. ranking of tokens) which all methods must uncover, and new feature attribution methods are evaluated based on how strongly they correlate with the ``ideal'' explanation. However, it is unclear whether this assumption holds. 
For instance, input token importance rankings may only capture a narrow slice of the model's behavior, such that many plausible rankings exist.  

Since many tasks may be too complex for humans to judge token-level importance, it can be unclear how to choose the ``ideal'' ranking. 
While a handful of highly polar tokens are generally indicative of the label in binary sentiment classification \cite{sun-lu-2020-understanding}, annotators may be unsure how to rank the other tokens. The difficulty increases in the pair-sequence setting -- if two words indicate a contradiction, which is more important?  
As a result, when agreement is measured in the presence of multiple faithful and plausible rankings, feature attribution methods may look deceptively problematic, even if they are not.

Figure~\ref{fig:distill-ps} shows an example of how different explanation methods can highlight different tokens as being important for the prediction, even though the underlying model is the same. 
Figure~\ref{fig:distill-ps} shows a textual entailment example from the MNLI dataset, where the task is to predict whether or not the second sentence entails the first. 
The first sentence comes after the [CLS] token: \textit{``But there are many more who still need our help.''}. 
The second sentence comes after the first [SEP] token: \textit{``10,000 people still need our help''}. 
Given that we see different explanations for the same example using the same model, it is unclear how to identify which of these explanations is the ``ideal'' one. 

Instead of evaluating potential explanations based on how well they correlate with one another, we suggest that researchers and practitioners should test various explanation methods and employ a human-in-the-loop process to determine if the explanations align with human intuition for the particular use case at hand. 
This can be used in combination with metrics such as those suggested by \citet{atanasova-etal-2020-diagnostic}. 

%!TEX root = ../main.tex

\section{Conclusion} 
\label{sec:conclusion}

In this work, we investigate how strongly attention-based explanations correlate with recent feature attribution methods for NLP tasks. 
We compare attention-based explanations to five recent feature attribution methods, using both transformer- and LSTM-based models, on both single- and pair-sequence tasks.
Overall, we observe a low degree of correlation between the attention-based explanations and the feature attribution explanations. 
For the transformer-based model, we find a weak correlation across all explanations for both task types. 
For the BiLSTM, we observe some strong correlations between explanations, but only for simple single-sequence tasks.

Through our experiments, we discover
\begin{enumerate*}[label=(\roman*)]
	\item a general lack of correlation between explanation methods, especially for more complex settings (i.e., transformer-based model and pair-sequence tasks) which is corroborated by additional recent research on text, tabular, and image data \cite{feldhus-etal-2021-thermostat,krishna_disagreement_2022},
	\item that similar explanations do not always result in correlated rankings, and 
	\item the existence of a single ``ideal'' explanation is questionable, which is a fundamental assumption of the \emph{agreement as evaluation} paradigm. 
\end{enumerate*}
Without an external ground-truth explanation, all that rank correlation tells us is whether or not two rankings are similar. 
For this reason, we recommend practitioners stop using \textit{agreement as evaluation} for attention-based explanations.  

In future work, we plan to approach this problem from first principles by formulating toy data to guarantee a single, correct top-$k$ or full ranking for each instance to see if XAI methods are consistently capable of recovering the ideal explanation under this setting.
% For future work, we plan to examine if our findings hold for other domains, such as image or video data. Attention has recently become more prominent in computer vision \cite{guo_attention_2022} and we believe it could be useful to compare various saliency map explanation methods to attention.

\section*{Acknowledgements}
We thank the anonymous reviewers, Bilal Alsallakh and Maarten de Rijke for their helpful discussions and suggestions.
This research was supported by Ahold Delhaize and the Netherlands Organisation for Scientific Research under project nr. 652.001.003 and the Nationale Politie.
All content represents the opinion of the authors, which is not necessarily shared or endorsed by their respective employers and/or sponsors.

\bibliographystyle{unsrtnat}
\bibliography{anthology,references}
\newpage
\appendix
\section{Model Performance}
\label{app:performance}

We include a uniform activation baseline to contextualize the attention mechanism's utility. Table \ref{tab:accuracy} notes the gap between the performance of uniform and softmax attention in the BiLSTM is never higher than 1-2\%. This distinction aligns with the results of \citet{wiegreffe-pinter-2019-attention} and \citet{vashishth2019attention}, who argue claims of attention-based interpretability are stronger in situations in which models need the attention module to solve the underlying task. Of course, it is difficult to prove a causal effect in a deep network: the BiLSTM may solve the task differently depending on whether or not the attention mechanism is ablated, meaning it is still possible to make claims of interpretability whether or not softmax attention leads to higher task performance. We again emphasize that we do not wish to take a side in the ``attention explanation" debate, but this point further stresses the difficulty of proving anything with the \emph{agreement as evaluation} paradigm.

\begin{table}[h!]
    \centering
    \caption{Test set accuracy using uniform and softmax activations in the attention mechanisms. A uniform activation of the attention weights makes the attention weights meaningless and, therefore, the drop in evaluation performance caused by using uniform attention weights gives an indication of the utility of the use attention layer(s) for each task.}
    \label{tab:accuracy}
    \begin{tabular}{crrrr}
        \toprule
                & \multicolumn{2}{c}{BiLSTM} & \multicolumn{2}{c}{DistillBERT} \\ 
          \cmidrule(r){2-3}\cmidrule(r){4-5}
         & Uniform & Softmax & Uniform & Softmax\\
        \midrule
        MNLI    & $.659 \pm .001$   & $.667 \pm .004$   & $.599 \pm .002$   & $.779 \pm .002$ \\
        Quora   & $.829 \pm .001$   & $.830 \pm .001$   & $.832 \pm .001$   & $.888 \pm .001$ \\
        SNLI    & $.804 \pm .004$   & $.807 \pm .002$   & $.770 \pm .005$   & $.871 \pm .001$ \\
        IMDb    & $.874 \pm .011$   & $.872 \pm .014$   & $.879 \pm .003$   & $.890 \pm .005$ \\
        SST-2   & $.823 \pm .008$   & $.826 \pm .011$   & $.823 \pm .004$   & $.842 \pm .003$ \\
    \bottomrule
    \end{tabular}
\end{table}

\section{Attention Flow}
\label{app:attention_flow}

Despite \emph{attention flow}, Grad-SHAP, and Deep-SHAP all (supposedly) being valid Shapley Value explanations \cite{ethayarajh-jurafsky-2021-attention}, Table \ref{tab:attn_flow_apx} shows that the agreement is low.

\begin{table*}[h!]
    \centering
    \caption{Mean Kendall-$\tau$ between the explanations given by \emph{attention flow} and our chosen XAI methods for the DistilBERT model when applied to 500 instances of the test portion of each dataset. IMDb is not included among these datasets, because the long sequences made the \emph{attention flow} computation unfeasible.}\label{tab:attn_flow_apx}
    \setlength\tabcolsep{2pt}
    \renewcommand{\arraystretch}{1.17}
    \resizebox{\textwidth}{!}{%
    \begin{tabular}{clRRRRR}
            &  & \multicolumn{1}{c}{LIME} & \multicolumn{1}{c}{Int-Grad} & \multicolumn{1}{c}{DeepLIFT} & \multicolumn{1}{c}{Grad-SHAP} & \multicolumn{1}{c}{Deep-SHAP}            \\
        \hhline{-------}\addlinespace[0.1mm]
        \multirow{4}{*}{\rot{Attn Flow}}
            & MNLI    & .1326  & .1251  & .2159  & .1227  & .2148   \\
            & Quora   & .0853  & .2426  & .0367  & .0241  & .2319   \\
            & SNLI    & .0844  & .0753  & .2178  & .0571  & .2149   \\
            & SST-2   & .1795  & .0689  & .1286  & .0811  & .1202   \\
        \hhline{-------}
    \end{tabular}}
\end{table*}\vspace{-1em}

\section{Result tables with standard deviation}
\label{app:results-with-std-dev}

In Tables \ref{app:distilbert} and \ref{app:BILSTM} we provide the same results as in Tables \ref{tab:distilbert_results} and \ref{tab:bilstm_results} of the main paper, but with the standard deviation for each experiment.

\begin{table*}[ht!]
	\centering
	\caption{Mean Kendall-$\tau$ plus standard deviation between the tested explanation methods for the DistillBERT. The single sequence datasets are indicated by using the \textit{italic} font type. Attn refers to Attention Rollout.}
	\label{app:BILSTM}
    \begin{tabular}{clcccccc}
            \toprule
             &  & \multicolumn{1}{c}{Attn}  & \multicolumn{1}{c}{LIME} & \multicolumn{1}{c}{Int-Grad} & \multicolumn{1}{c}{DeepLIFT} & \multicolumn{1}{c}{Grad-SHAP}  & \multicolumn{1}{c}{Deep-SHAP} \\
            \midrule
            \multirow{5}{*}{\rot{Attn}}
            	& \textit{IMDb} & 1.  $\pm$ .0  & .1259 $\pm$ .1000  & .1818 $\pm$ .1256  & .2516 $\pm$ .0752 & .1432 $\pm$ .1296 & .2303 $\pm$ .0826 \\
            	& \textit{SST-2} & 1.  $\pm$ .0& .1359 $\pm$ .1772  & .0511 $\pm$ .1680  & .1328 $\pm$ .1764 & .0737 $\pm$ .1629 & .1291 $\pm$ .1788 \\
                & MNLI  & 1.  $\pm$ .0 & .2678 $\pm$ .1196 & .1891 $\pm$ .1302  & .2432 $\pm$ .1253 & .1905 $\pm$ .1372 & .2067 $\pm$ .1609 \\
                & Quora & 1.  $\pm$ .0 & .1622 $\pm$ .1419 & .0574 $\pm$ .1640  & .2267 $\pm$ .1374 & .0518 $\pm$ .1652 & .2257 $\pm$ .1383 \\
                & SNLI   & 1.  $\pm$ .0& .1434 $\pm$ .1649  & .1645 $\pm$ .1798  & .2214 $\pm$ .1431 & .1600 $\pm$ .1720 & .1796 $\pm$ .2004 \\
            \midrule
            \multirow{5}{*}{\rot{Lime}}
            	& \textit{IMDb}   & & 1.  $\pm$ .0 & .1050 $\pm$ .1069  & .0696 $\pm$ .0791 & .0929 $\pm$ .0983 & .0655 $\pm$ .0796 \\
            	& \textit{SST-2}  & & 1.  $\pm$ .0 & .2861 $\pm$ .1658  & .0618 $\pm$ .1702 & .2414 $\pm$ .1715 & .0499 $\pm$ .1668 \\
                & MNLI   & & 1.  $\pm$ .0 & .1794 $\pm$ .1324  & .1526 $\pm$ .1291 & .1592 $\pm$ .1367 & .1205 $\pm$ .1493 \\
                & Quora  & & 1.  $\pm$ .0 & .1407 $\pm$ .1632  & .0032 $\pm$ .1555 & .1144 $\pm$ .1597 & .0095 $\pm$ .1550 \\
                & SNLI   && 1.  $\pm$ .0  & .1529 $\pm$ .1588  & .0925 $\pm$ .1645 & .1104 $\pm$ .1596 & .0593 $\pm$ .1710 \\
           \midrule
            \multirow{5}{*}{\rot{Int-Grad}}
            	& \textit{IMDb}   &  & & 1.  $\pm$ .0 & .1433 $\pm$ .1443 & .5495 $\pm$ .2340 & .1246 $\pm$ .1335 \\
            	& \textit{SST-2}  &  & & 1.  $\pm$ .0 & .0498 $\pm$ .1897 & .4987 $\pm$ .2405 & .0381 $\pm$ .1885 \\
                & MNLI   &  && 1.  $\pm$ .0  & .2153 $\pm$ .1748 & .4780 $\pm$ .2441 & .1708 $\pm$ .1732 \\
                & Quora  &  & & 1.  $\pm$ .0 & .0625 $\pm$ .2088 & .4674 $\pm$ .2930 & .0529 $\pm$ .1969 \\
                & SNLI   &  & & 1.  $\pm$ .0 & .0955 $\pm$ .1801 & .3932 $\pm$ .2588 & .0700 $\pm$ .1829 \\
             \midrule
            \multirow{5}{*}{\rot{DeepLIFT}}
            	& \textit{IMDb}   &  &  & & 1.  $\pm$ .0 & .1306 $\pm$ .1532 & .4830 $\pm$ .2469 \\
            	& \textit{SST-2}  &  &  & & 1.  $\pm$ .0 & .0522 $\pm$ .2059 & .4514 $\pm$ .3031 \\
                & MNLI   &  &  & & 1.  $\pm$ .0 & .2324 $\pm$ .2078 & .4985 $\pm$ .2949 \\     
                & Quora  &  &  & & 1.  $\pm$ .0 & .0637 $\pm$ .2272 & .5951 $\pm$ .3210 \\
                & SNLI   &  &  && 1.  $\pm$ .0 & .1181 $\pm$ .2120 & .5554 $\pm$ .3576 \\
            \midrule
            \multirow{5}{*}{\rot{Grad-SHAP}}
	            & \textit{IMDb}  &  &  &  & & 1.  $\pm$ .0 & .1093 $\pm$ .1342 \\
	            & \textit{SST-2} &  &  &  & & 1.  $\pm$ .0 & .0419 $\pm$ .1919 \\
                & MNLI  &  &  &  & & 1.  $\pm$ .0 & .1752 $\pm$ .1902 \\
                & Quora &  &  &  & & 1.  $\pm$ .0 & .0535 $\pm$ .2131 \\
                & SNLI  &  &  & & & 1.  $\pm$ .0  & .0851 $\pm$ .2108 \\
            \midrule
	            \multirow{5}{*}{\rot{Deep-SHAP}}
	            & \textit{IMDb} &    &        &         &     &     & 1.  $\pm$ .0   \\
	            & \textit{SST-2} &  &        &         &     &      & 1.  $\pm$ .0  \\
	            & MNLI  &  &        &         &     &      &  1.  $\pm$ .0   \\
	            & Quora &  &        &         &     &       & 1.  $\pm$ .0   \\
	            & SNLI &   &        &         &     &       &  1.  $\pm$ .0 \\   
            \bottomrule
    \end{tabular}
\end{table*}

\begin{table*}[ht!]
	\centering
	\caption{Mean Kendall-$\tau$ plus standard deviation between the tested explanation methods for the BiLSTM. The single sequence datasets are indicated by using the \textit{italic} font type. Attn refers to Attention Weights.}
	\label{app:distilbert}
	\begin{tabular}{clcccccc}
            \toprule
            &  & \multicolumn{1}{c}{Attn} & \multicolumn{1}{c}{LIME} & \multicolumn{1}{c}{Int-Grad} & \multicolumn{1}{c}{DeepLIFT} & \multicolumn{1}{c}{Grad-SHAP}  & \multicolumn{1}{c}{Deep-SHAP} \\
            \midrule
            \multirow{5}{*}{\rot{Attn}}
	            & \textit{IMDb} & 1.  $\pm$ .0  & .2014 $\pm$ .0790  & .2188 $\pm$ .0815  & .2494 $\pm$ .1028 & .2209 $\pm$ .0834 & .2309 $\pm$ .1219 \\
	            & \textit{SST-2} & 1.  $\pm$ .0 & .1326 $\pm$ .2372  & .1093 $\pm$ .2554  & .1372 $\pm$ .2575 & .1101 $\pm$ .2534 & .1400 $\pm$ .2672 \\
	            & MNLI  & 1.  $\pm$ .0   & .1958 $\pm$ .1496 & .2523 $\pm$ .1654  & .2549 $\pm$ .1570 & .2473 $\pm$ .1651 & .2370 $\pm$ .1550 \\
	            & Quora  & 1.  $\pm$ .0  & .0363 $\pm$ .2243 & .0143 $\pm$ .2019  & .0894 $\pm$ .2084 & .0182 $\pm$ .2060 & .1017 $\pm$ .2311 \\
	            & SNLI   & 1.  $\pm$ .0   & .2198 $\pm$ .1761  & .2566 $\pm$ .1748  & .3158 $\pm$ .1555 & .2517 $\pm$ .1782 & .2938 $\pm$ .1556 \\
            \midrule
            \multirow{5}{*}{\rot{Lime}}
	            & \textit{IMDb}   &  & 1.  $\pm$ .0  & .6538 $\pm$ .1557  & .5854 $\pm$ .1495 & .6486 $\pm$ .1544 & .5584 $\pm$ .1609 \\
	            & \textit{SST-2}  &  &     1.  $\pm$ .0 & .4968 $\pm$ .2739  & .4734 $\pm$ .2669 & .4962 $\pm$ .2751 & .4422 $\pm$ .2704 \\
	            & MNLI   &  &    1.  $\pm$ .0 & .3281 $\pm$ .1417  & .2444 $\pm$ .1452 & .3187 $\pm$ .1398 & .2269 $\pm$ .1495 \\
	            & Quora  &  &    1.  $\pm$ .0& .2099 $\pm$ .1943  & .1900 $\pm$ .1887 & .2037 $\pm$ .1939 & .1670 $\pm$ .2036 \\
	            & SNLI   &  &    1.  $\pm$ .0 & .2673 $\pm$ .1650  & .1676 $\pm$ .1640 & .2481 $\pm$ .1688 & .1566 $\pm$ .1682 \\
        	\midrule
            \multirow{5}{*}{\rot{Int-Grad}}
	           	& \textit{IMDb}   &  & & 1.  $\pm$ .0 & .7331 $\pm$ .1155 & .9409 $\pm$ .0504 & .6994 $\pm$ .1342 \\
	           	& \textit{SST-2}  &  &  & 1.  $\pm$ .0 & .8683 $\pm$ .1032 & .9707 $\pm$ .0382 & .8063 $\pm$ .1561 \\
	            & MNLI   &  &  & 1.  $\pm$ .0 & .4984 $\pm$ .1561 & .8138 $\pm$ .1219 & .4021 $\pm$ .1690 \\
	            & Quora  &  & & 1.  $\pm$ .0 & .2906 $\pm$ .2374 & .7420 $\pm$ .2089 & .2290 $\pm$ .2343 \\
	            & SNLI   &  & & 1.  $\pm$ .0  & .2461 $\pm$ .1738 & .6535 $\pm$ .1848 & .2165 $\pm$ .1800 \\
            \midrule
            \multirow{5}{*}{\rot{DeepLIFT}}
	            & \textit{IMDb}   &  &  & & 1.  $\pm$ .0  & .7378 $\pm$ .1192 & .8593 $\pm$ .1453 \\
	            & \textit{SST-2}  &  &  & & 1.  $\pm$ .0  & .8682 $\pm$ .1068 & .8729 $\pm$ .1442 \\
	            & MNLI   &  &  & & 1.  $\pm$ .0  & .4987 $\pm$ .1732 & .6208 $\pm$ .2175 \\     
	            & Quora  &  &  & & 1.  $\pm$ .0  & .3158 $\pm$ .2473 & .6179 $\pm$ .3241 \\
	            & SNLI   &  &  & & 1.  $\pm$ .0   & .2557 $\pm$ .1937 & .5791 $\pm$ .2748 \\
            \midrule
            \multirow{5}{*}{\rot{Grad-SHAP}}
	           	& \textit{IMDb}  &  &  &  & & 1.  $\pm$ .0  & .7021 $\pm$ .1366 \\
	           	& \textit{SST-2} &  &  &  &  & 1.  $\pm$ .0  & .8056 $\pm$ .1566 \\
	            & MNLI  &  &  &  &  & 1.  $\pm$ .0  & .4015 $\pm$ .1757 \\
	            & Quora &  &  &  &  & 1.  $\pm$ .0  & .2433 $\pm$ .2417 \\
	            & SNLI  &  &  & &  & 1.  $\pm$ .0  & .2219 $\pm$ .1927 \\
           \midrule
           \multirow{5}{*}{\rot{Deep-SHAP}}
	           & \textit{IMDb} &    &        &         &     &     & 1.  $\pm$ .0   \\
	           & \textit{SST-2} &  &        &         &     &      & 1.  $\pm$ .0  \\
	           & MNLI  &  &        &         &     &      &  1.  $\pm$ .0   \\
	           & Quora &  &        &         &     &       & 1.  $\pm$ .0   \\
	           & SNLI &   &        &         &     &       &  1.  $\pm$ .0 \\   
           \bottomrule
    \end{tabular}
\end{table*}

\section{Reproducibility}
\label{app:reproducibility}

Our code is publicly available at \url{https://github.com/sfschouten/court-of-xai}.
We conducted our experiments on Amazon Web Services \texttt{g4dn.xlarge} EC2 instances using an NVIDIA T4 GPU with 16GB of RAM. The version of PyTorch was \texttt{1.6.0+cu101}.
We refer to Table \ref{tab:runtime} for the average time to train each model on each dataset.

The DistilBERT model contained 66955779 trainable parameters and the BiLSTM model contained 12553519 trainable parameters, as reported by the AllenNLP library \cite{gardner-etal-2018-allennlp}.
Table \ref{tab:dataset_statistics} lists the number of instances in each split of each dataset and Table \ref{tab:val_accuracy} details the accuracy of our models on the validation sets during training.

Links to download versions of all datasets are included in our code repository. For posterity, links to all datasets are listed here: 
\begin{itemize}
	\item \textbf{SST-2}: \url{https://github.com/successar/AttentionExplanation/tree/master/preprocess/SST}
	\item \textbf{IMDb}: \url{https://github.com/successar/AttentionExplanation/tree/master/preprocess/IMDB}
	\item \textbf{SNLI}: \url{https://nlp.stanford.edu/projects/snli/}
	\item \textbf{MNLI}: \url{https://cims.nyu.edu/~sbowman/multinli/}
	\item \textbf{XNLI}: \url{https://cims.nyu.edu/~sbowman/xnli/}
	\item \textbf{Quora Question Pair}: \url{https://drive.google.com/file/d/12b-cq6D45U5c-McPoq2wsFjzs6QduY_y/view?usp=sharing}
\end{itemize}

\begin{table}[h!]
    \centering
    \caption{Number of minutes (average $\pm$ standard deviation) required to train each model on each dataset reported across three seeds.}
    \label{tab:runtime}
    \begin{tabular}{crr}
     \toprule
        & BiLSTM & DistilBERT \\
        \midrule
        MNLI    & 8.65 $\pm$ 0.635 & 296.228 $\pm$ 48.859 \\
        Quora   & 7.567 $\pm$ 1.404 & 380.056 $\pm$ 124.911 \\
        SNLI    & 31.495 $\pm$ 5.618 & 126.395 $\pm$ 22.909 \\
        IMDb    & 1.122 $\pm$ 0.107 & 24.2 $\pm$ 1.212 \\
        SST-2   & 0.216 $\pm$ 0.029 & 2.833 $\pm$ 0.65 \\
    \bottomrule
    \end{tabular}
\end{table}

\begin{table}[h!]
    \centering
    \caption{Number of instances in each split of each dataset before any exclusions based on length (see Section 4.1). Since MultiNLI has no publicly available test set, we use the English subset of the XNLI dataset.}
    \label{tab:dataset_statistics}
    \begin{tabular}{crrr}
     \toprule
        & Training & Validation & Test \\
        \midrule
        MNLI    & 392702 & 10000 & 5000 \\
        Quora   & 323426 & 40429 & 40431 \\
        SNLI    & 550152 & 10000 & 10000 \\
        IMDb    & 17212 & 4304 & 4363 \\
        SST-2   & 8544 & 1101 & 2210 \\
    \bottomrule
    \end{tabular}
\end{table}

\begin{table}[h!]
    \centering
    \caption{Validation accuracy (average $\pm$ standard deviation) of the selected model epoch reported across three seeds.}
    \label{tab:val_accuracy}
    \begin{tabular}{ccc}
     \toprule
        & BiLSTM & DistilBERT \\
        \midrule
        MNLI    & 67.088 $\pm$ 0.190 & 77.338 $\pm$ 0.251 \\
        Quora   & 83.232 $\pm$ 0.139 & 88.801 $\pm$ 0.055 \\
        SNLI    & 81.535 $\pm$ 0.041 & 87.679 $\pm$ 0.075 \\
        IMDb    & 87.975 $\pm$ 1.375 & 88.587 $\pm$ 0.489 \\
        SST-2   & 80.696 $\pm$ 0.403 & 83.066 $\pm$ 0.692 \\
    \bottomrule
    \end{tabular}
\end{table}

\end{document}